\def\year{2019}\relax
\documentclass[letterpaper]{article}
\usepackage{aaai19}
\usepackage{times}
\usepackage{helvet}
\usepackage{courier}
\usepackage{url}
\usepackage{graphicx}
\usepackage{subfig}
\usepackage{graphicx}
\usepackage{amsfonts,amssymb}
\usepackage{amsmath}
\usepackage{array}
\usepackage{verbatim}

\begin{document}
\title{Deep Short Text Classification with Knowledge Powered Attention}
\author{Jindong Chen\textsuperscript{\rm 12}, 
	Yizhou Hu\textsuperscript{\rm 1}, 
	Jingping Liu\textsuperscript{\rm 1}, 
	Yanghua Xiao\textsuperscript{\rm 1345*}, 
	Haiyun Jiang\textsuperscript{\rm 1} \\
\textsuperscript{\rm 1}Shanghai Key Laboratory of Data Science, School of Computer Science, Fudan University, China\\
\textsuperscript{\rm 2}CETC Big Data Research Institute Co.,Ltd., Guizhou, China\\
\textsuperscript{\rm 3}Shanghai Institute of Intelligent Electronics \& Systems, Shanghai, China\\
\textsuperscript{\rm 4}Shuyan Technology, Shanghai, China\\
\textsuperscript{\rm 5}Alibaba Group, Zhejiang, China\\
\{chenjd16, huyz15, jpliu17, shawyh, jianghy16\}@fudan.edu.cn\\
}

\maketitle
\begin{abstract}
	Short text classification is one of important tasks in Natural Language Processing (NLP).
	Unlike paragraphs or documents, short texts are more ambiguous since they have not enough contextual information, which poses a great challenge for classification.
	In this paper, we retrieve knowledge from external knowledge source
	to enhance the semantic representation of short texts. 
	We take conceptual information as a kind of knowledge and incorporate it into deep neural networks.
	For the purpose of measuring the importance of knowledge, we introduce attention mechanisms and propose deep Short Text Classification with Knowledge powered Attention (STCKA).
	We utilize Concept towards Short Text (C-ST) attention and Concept towards Concept Set (C-CS) attention to acquire the weight of concepts from two aspects. 
	And we classify a short text with the help of conceptual information.
	Unlike traditional approaches, our model acts like a human being who has intrinsic ability to make decisions based on observation (i.e., training data for machines) and pays more attention to important knowledge.
	We also conduct extensive experiments on four public datasets for different tasks. The experimental results and case studies show that our model outperforms the state-of-the-art methods, justifying the effectiveness of knowledge powered attention.
\end{abstract}

\section{Introduction}
Short text classification is one of important ways to understand short texts and is useful in a wide range of applications including sentiment analysis \cite{Wang2014Concept}, dialog system \cite{Lee2016Sequential} and user intent understanding \cite{hu2009understanding}. Compared with paragraphs or documents, short texts are more \textit{ambiguous} since they have not enough contextual information, which poses a great challenge for short text classification.
Existing methods \cite{Gabrilovich2007Computing,Wang2014Concept,Wang2014Concept} for short text classification can be mainly divided into two categories: \textit{explicit representation} and \textit{implicit representation} \cite{understanding-short-texts}.

For explicit representation, a short text is represented as a sparse vector where each dimension is an explicit feature, corresponding to syntactic information of the short text including n-gram, POS tagging and syntactic parsing \cite{Pang2002Thumbs}. Researchers develop effective features from many different aspects such as knowledge base and the results of dependency parsing. The explicit model is interpretable and easy to understand for human beings. 
However, the explicit representation usually ignores the context of short text and cannot capture deep semantic information.

In terms of implicit representation, a short text is usually mapped to an implicit space and represented as a dense vector \cite{Mikolov2013Efficient}. 
The implicit model is good at capturing syntax and semantic information in short text based on deep neural networks. However, it ignores important semantic relations such as \textit{isA} and \textit{isPropertyOf} that exist in Knowledge Bases (KBs). Such information is helpful for the understanding of short texts, especially when dealing with previously unseen words. 
For example, given a short text S1: ``\textit{Jay grew up in Taiwan}'', the implicit model may treat \texttt{Jay} as a new word and cannot capture that \texttt{Jay} is a \texttt{singer} which is beneficial to classify the short text into the class \textit{entertainment}.

In this paper, we integrate explicit and implicit representation of short texts into a unified deep neural network model. 
We enrich the semantic representation of short texts with the help of explicit KBs such as YAGO \cite{Suchanek2008Yago} and Freebase \cite{bollacker2008freebase}. 
This allows the model to retrieve knowledge from an external knowledge source that is not explicitly stated in the short text but relevant for classification.
As the example shown in S1, the conceptual information as a kind of knowledge is helpful for classification.
Therefore, we utilize \textit{isA} relation and associate each short text with its relevant concepts in KB by conceptualization\footnote{Conceptualization refers to the process of retrieving the conceptual information of short text from KBs.}.
Afterwards we incorporate the \textit{conceptual information} as \textit{prior knowledge} into deep neural networks.

Although it may seem intuitive to simply integrate conceptual information into a deep neural network, there are still two major problems. 
First, when conceptualizing the short text, some improper concepts are easily introduced due to the ambiguity of entities or the noise in KBs.
For example, in the short text S2: ``\textit{Alice has been using Apple for more than 10 years}'', we acquire the concepts \texttt{fruit} and \texttt{mobile phone} of \texttt{apple} from KB. Obviously, \texttt{fruit} is not an appropriate concept here which is caused by the ambiguity of \texttt{apple}.
Second, it is necessary to take into account the granularity of concepts and the relative importance of the concepts.
For instance, in the short text S3: ``\textit{Bill Gates is one of the co-founders of Microsoft}'', we retrieve the concepts \texttt{person} and \texttt{entrepreneur} of \texttt{Bill Gates} from KB. Although they are both correct concepts, \texttt{entrepreneur} is more specific than \texttt{person} and should be assigned a larger weight in such a scenario. 
Prior work \cite{Gabrilovich2007Computing,Wang2017Combining} exploited web-scale KBs for enriching the short text representation, but without carefully addressing these two problems.

To solve the two problems, we introduce \textit{attention mechanisms} and propose deep \textbf{S}hort \textbf{T}ext \textbf{C}lassification with \textbf{K}nowledge Powered \textbf{A}ttention (STCKA). Attention mechanism has been widely used to acquire the weight of vectors in many NLP applications including machine translation \cite{Bahdanau2015Neural}, abstractive summarization \cite{Zeng2016Efficient} and question answering \cite{Hao2017An}.
For the first problem, we use \textit{Concept towards Short Text (C-ST) attention} to measure the semantic similarity between a short text and its corresponding concepts.
Our model assigns a larger weight to the concept \texttt{mobile phone} in S2 since it is more semantically similar to the short text than the concept \texttt{fruit}.
For the second problem, we use \textit{Concept towards Concept Set (C-CS) attention} to explore the importance of each concept with respect to the whole concept set. Our model assigns a larger weight to the concept \texttt{entrepreneur} in S3 which is more discriminative for a specific classification task. 

We introduce a \textit{soft switch} to combine two attention weights into one and produce the final attention weight of each concept, which is adaptively learned by our model on different datasets.
Then we calculate a weighted sum of the concept vectors to produce the concept representation. 
Besides, we make full use of both character and word level features of short texts and employ self-attention to generate the short text representation. Finally, we classify a short text based on the representation of short text and its concepts.
The main contributions of this paper are summarized as follows:
\begin{itemize}
	\item We propose deep Short Text Classification with Knowledge Powered Attention. As far as we know, this is the first attention model which combines prior knowledge in KBs to enrich the semantic information of the short text. 
	\item We introduce two attention mechanisms (i.e., C-ST and C-CS attention) to measure the importance of each concept from two aspects and combine them by a soft switch to acquire the weight of concept adaptively.
	\item We conduct extensive experiments on four datasets for different tasks. The results show that our model outperforms the state-of-the-art methods.
	
\end{itemize}

\begin{figure*}[!hbt]
	\centering
	\includegraphics[width=0.8\linewidth]{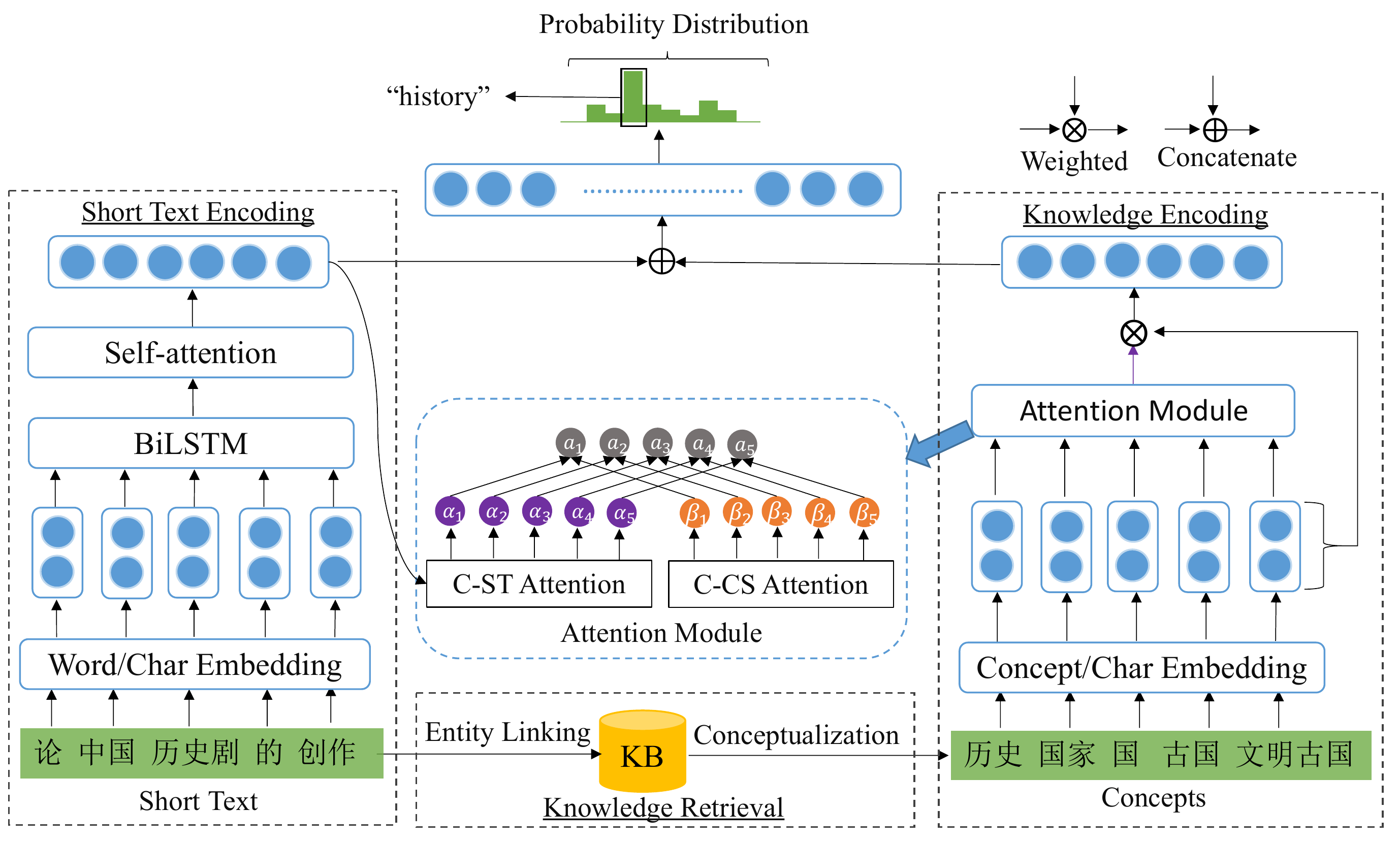}
	\caption{Model architecture. The input short text is \textit{on the creation of Chinese historical plays}. The concepts include \textit{history, country}, etc. The class label is \textit{history}.}
	\label{fig:model}
\end{figure*}

\section{Our Model}
Our model STCKA is a knowledge-enhanced deep neural network shown in Figure \ref{fig:model}.
We provide a brief overview of our model before detailing it. 
The input of the network is a short text $s$, which is a sequence of words.
The output of the network is the probability distribution of class labels.
We use $p(y|s,\phi)$ to denote the probability of a short text being class $y$, where $\phi$ is the parameters in the network.
Our model contains four modules. \textit{Knowledge Retrieval} module retrieves conceptual information relevant to the short text from KBs.
\textit{Input Embedding} module utilizes the character and word level features of the short text to produce the representation of words and concepts. 
\textit{Short Text Encoding} module encodes the short text by self-attention and produces short text representation $q$. \textit{Knowledge Encoding} module applies two attention mechanisms on concept vectors to obtain the concept representation $p$.
Next, we concatenate $p$ and $q$ to fuse the short text and conceptual information, which is fed into a fully connected layer. Finally, we use an output layer to acquire the probability of each class label. 

\subsection{Knowledge Retrieval}\label{stc}
The goal of this module is to retrieve relevant knowledge from KBs. 
This paper takes \textit{isA} relation as an example, and other semantic relations such as \textit{isPropertyOf} can also be applied in a similar way.
Specifically, given a short text $ s $, we hope to find a concept set $\mathcal C$ relevant to it. We achieve this goal by two major steps: \textit{entity linking} and \textit{conceptualization}. Entity linking is an important task in NLP and is used to identify the entities mentioned in the short text \cite{moro2014entity}. We acquire an entity set $\mathcal E$ of a short text by leveraging the existing entity linking solutions \cite{chen2018short}. Then, for each entity $ e \in \mathcal E $, we acquire its conceptual information from an existing KB, such as YAGO \cite{Suchanek2008Yago}, Probase \cite{Wu2012Probase} and CN-Probase \cite{cnprobase} by conceptualization.
For instance, given a short text ``\textit{Jay and Jolin are born in Taiwan}'',  we obtain the entity set $\mathcal E$ = \{\texttt{Jay Chou}, \texttt{Jolin Tsai}\} by entity linking. 
Then, we conceptualize the entity \texttt{Jay Chou} and acquire its concept set $\mathcal C$ = \{\texttt{person}, \texttt{singer}, \texttt{actor}, \texttt{musician}, \texttt{director}\} from CN-Probase.

\subsection{Input Embedding}\label{ie}
The input consists of two parts: short text $ s $ of length $n$ and concept set $\mathcal C$ of size $m$.
We use three kinds of embeddings in this module including character embedding, word embedding, and concept embedding.
\textit{Character embedding layer} is responsible for mapping each
word to a high-dimensional vector space. We obtain the character level embedding of each word using Convolutional Neural Networks (CNN). Characters are embedded into vectors, which can be considered as 1D inputs to the CNN, and whose size is the input channel size of the CNN. The outputs of the CNN are max-pooled over the entire width to obtain a fixed-size vector for each word. 
\textit{Word and concept embedding layer} also maps each word and concept to a high-dimensional vector space. We use pre-trained word vectors \cite{Mikolov2013Efficient} to obtain the word embedding of each word. The dimension of word vectors, character vectors and concept vectors is $\frac{d}{2}$.
We concatenate the character embedding vectors and word/concept embedding vectors to obtain $d$-dimensional word/concept representation.

\subsection{Short Text Encoding}\label{str}
The goal of this module is to produce the short text representation $q$ for a given short text of length $ n $ which is represented as the sequence of $d$-dimensional word vectors $ (x_1, x_2, ..., x_n) $. Self-attention is a special case of attention mechanism that only requires a single sequence to compute its representation \cite{Vaswani2017Attention}. Before using self-attention, we add a recurrent neural network (RNN) to transform the inputs from the bottom layers. The reason is explained as follows. Attention mechanism uses weighted sum to generate output vectors, thus its representational power is limited. Meanwhile, RNN is good at capturing the contextual information of sequence, which can further increase the expressive power of attentional network.

In this paper, we employ bidirectional LSTM (BiLSTM) as \cite{Hao2017An} does, which consists of both forward and backward networks to process the short text:
\begin{equation}
\overrightarrow{h_t}=\overrightarrow{LSTM}(x_t, \overrightarrow{h_{t-1}})
\end{equation}
\begin{equation}
\overleftarrow{h_t}=\overleftarrow{LSTM}(x_t, \overleftarrow{h_{t+1}})
\end{equation}
We concatenate each $ \overrightarrow{h_t} $ and $ \overleftarrow{h_t} $ to obtain a hidden state $ h_t $. 
Let the hidden unit number for each unidirectional LSTM be $ u $.
For simplicity,  we denote all the $ h_t $s as $ H \in \mathbb{R}^{n\times2u} $:
\begin{equation}
H=(h_1, h_2,...h_n)
\end{equation}

Then, we use the scaled dot-product attention, which is a variant of dot-product (multiplicative) attention \cite{luong2015effective}. The purpose is to learn the word dependence within the sentence and capture the internal structure of the sentence.
Given a matrix of $n$ query vectors  $ Q \in \mathbb{R}^{n\times2u} $, keys $ K \in \mathbb{R}^{n\times2u} $ and values $ V \in \mathbb{R}^{n\times2u} $, the scaled dot-product attention computes the attention scores based on the following equation:
\begin{equation}\label{key}
A=Attention(Q,K,V)=softmax(\frac{QK^T}{\sqrt{2u}})V
\end{equation}
Here $Q,K,V$ are the same matrix and equal to $H$, $\frac{1}{\sqrt{2u}}$ is the scaling factor. The output of attention layer is a matrix denoted as $A \in\mathbb{R}^{n\times 2u}$. 
Next, we use max-pooling layer over $A$ to acquire the short text representation $ q\in \mathbb{R}^{2u}$. The idea is to choose the highest value on each dimension of vector to capture the most important feature.

\subsection{Knowledge Encoding}\label{am}
The prior knowledge obtained from external resources such as knowledge bases provides richer information to help decide the class label given a short text. 
We take conceptual information as an example to illustrate knowledge encoding, and other prior knowledge can also be used in a similar way.
Given a concept set $\mathcal C$ of size $m$ denoted as $ (c_1, c_2, ..., c_m) $ where $c_i$ is the $i$-th concept vector, we aim at producing its vector representation $p$. We first introduce two attention mechanisms to pay more attention to important concepts.

To reduce the bad influence of some improper concepts introduced due to the ambiguity of entities or the noise in KBs, we propose \textit{Concept towards Short Text (C-ST) attention} based on vanilla attention \cite{Bahdanau2015Neural} to measure the semantic similarity between the $i$-th concept and short text representation $q$.
We use the following formula to calculate the C-ST attention:
\begin{equation}
\alpha_i = softmax(w_1^Tf(W_1[c_i;q]+b_1))
\end{equation}
Here $\alpha_i$ denotes the weight of attention from $i$-th concept towards the short text. A larger $ \alpha_i $ means that the $ i$-th concept is more semantically similar to the short text. $f(\cdot)$ is a non-linear activation function such as hyperbolic tangent transformation and $ softmax $ is used to normalize attention weight of each concept. $ W_1 \in \mathbb{R}^{d_a\times (2u+d)} $ is a weight matrix and $ w_1 \in \mathbb{R}^{d_a} $ is a weight vector where $ d_a $ is a hyper-parameter, and $b_1$ is the offset.

Besides, in order to take the relative importance of the concepts into consideration, we propose \textit{Concept towards Concept Set (C-CS) attention} based on \textit{source2token self-attention} \cite{Lin2017A} to measure the importance of each concept with respect to the whole concept set.
We define the C-CS attention of each concept as follows:
\begin{equation}
\beta_i=softmax(w_2^Tf(W_2c_i)+b_2)
\end{equation}
Here $\beta_i$ denotes the weight of attention from the $i$-th concept towards whole concept set. $ W_2\in \mathbb{R}^{d_b\times d} $ is a weight matrix and $ w_2 \in \mathbb{R}^{d_b} $ is a weight vector where $ d_b $ is a hyper-parameter, and $b_2$ is the offset. 
The effect of C-CS attention is similar to that of \textit{feature selectio}n. It is a ``soft'' feature selection which assigns a larger weight to a vital concept, and a small weight (close to zero) to a trivial concept. More details are given in the experimental Section ``Knowledge Attention''. 

We combine $ \alpha_i $ and $ \beta_i $ by the following formula to obtain the final attention weight of each concept:
\begin{equation}\label{equ:attention}
\begin{split}
a_i&=softmax(\gamma\alpha_i+(1-\gamma)\beta_i) \\
&=\frac{exp(\gamma\alpha_i+(1-\gamma)\beta_i)}{\sum_{k\in [1,m]}^{}exp(\gamma\alpha_k+(1-\gamma)\beta_k)}
\end{split}
\end{equation}
Here $a_i$ denotes the final attention weight from the $i$-th concept towards the short text, $ \gamma \in[0,1]$ is a \textit{soft switch} to adjust the importance of two attention weights $ \alpha_i $ and $ \beta_i $.  
There are various ways to set the parameter $\gamma$.
The simplest one is to treat $\gamma$ as a hyper-parameter and manually adjust to obtain the best performance. Alternatively, $\gamma$ can also be learned by a neural network automatically. We choose the latter approach since it adaptively assigns different values to $\gamma$ on different datasets and achieves better experimental results.
We calculate $\gamma$ by the following formula:
\begin{equation}
\gamma = \sigma(w^T[\alpha;\beta]+b)
\end{equation}
where vectors $w$ and scalar $b$ are learnable parameters and $\sigma$ is the sigmoid function.
In the end, the final attention weights are employed to calculate a weighted sum of the concept vectors, resulting in a semantic vector that represents the concepts:
\begin{equation}
p=\sum_{i=1}^{m}a_i{c_i}
\end{equation}

\subsection{Training}
To train the model, we denote all the parameters to be trained as a set $ \phi $.
The training target of the network is used to maximize the log-likelihood with respect to $\phi$:
\begin{equation}
\phi \longmapsto \sum\limits_{s \in S}log \ p(y|s, \phi)
\end{equation}
where $S$ is the training short text set and $y$ is the correct class of short text $s$.

\section{Experiment}

\begin{table*}[!hbt]
	\center
	\begin{tabular}{|p{2.3cm}|p{1.1cm}|p{4cm}|p{1.6cm}|p{1.8cm}|p{1.3cm}|p{1.4cm}|}
		\hline
		Datasets&\# Class&Training/Validation/Test set& Avg. Chars & Avg. Words & Avg. Ent & Avg. Con\\
		\hline
		Weibo&7&3771/665/500&26.51&17.23&1.35&3.01 \\ 
		\hline
		Product Review&2&7648/1350/1000&64.71&40.31&1.82&4.87 \\ 
		\hline
		News Title&18&154999/27300/10000&20.63&12.02&1.35&2.72 \\ 
		\hline
		Topic&20&6170/1090/700&15.64&7.99&1.77&4.50 \\ 
		\hline
	\end{tabular}
	\caption{Details of the experimental datasets.}
	\label{table:datasets}
\end{table*}

\begin{table*}[!hbt]
	\center
	\begin{tabular}{|p{3cm}|p{1.5cm}|p{1.5cm}|p{2.5cm}|p{2cm}|}
		\hline
		Model&Weibo&Topic&Product Review&News Title\\
		\hline
		CNN&0.3900&0.8243&0.7290&0.7706 \\
		\hline
		RCNN&0.4040&0.8257&0.7280&0.7853 \\
		\hline
		CharCNN&0.4100&0.8500&0.7010&0.7493 \\
		\hline
		BiLSTM-MP&0.4160&0.8186&0.7290&0.7719 \\
		\hline
		BiLSTM-SA&0.4120&0.8200&0.7310&0.7802 \\
		\hline
		KPCNN&0.4240&0.8643&0.7340&0.7878 \\
		\hline
		STCKA&\textbf{0.4320}&\textbf{0.8814}&\textbf{0.7430}&\textbf{0.8011} \\
		\hline
	\end{tabular}
	\caption{Accuracy of compared models on different datasets.}
	\label{table:models}
\end{table*}

\subsection{Dataset}
We conduct experiments on four datasets, as shown in Table \ref{table:datasets}.
The first one is a \textit{Chinese Weibo emotion analysis}\footnote{http://tcci.ccf.org.cn/conference/2013/pages/page04\_sam.html} dataset from NLPCC2013 \cite{Zhou2017Emotional}. There are 7 kinds of emotions in these weibos, such as anger, disgust, fear and etc. 
The second one is \textit{product review}\footnote{http://tcci.ccf.org.cn/conference/2014/pages/page04\_sam.html} dataset from NLPCC2014  \cite{Zhou2017A}. The polarity of each review is binary, either positive or negative.
The third one is the \textit{Chinese news title}\footnote{http://tcci.ccf.org.cn/conference/2017/taskdata.php} dataset with 18 classes (e.g., entertainment, game, food) from NLPCC2017 \cite{qiu2017overview}.

The average word length of the above-mentioned three datasets is over 12.
To test whether our model works on much shorter texts, we build the \textit{Topic} dataset whose average word length is 7.99. The Topic dataset is collected from Sogou news \cite{fu2015name} where each news contains a title, document and topic (e.g., military, politics). We acquire the title as short text and topic as label. 
Besides, we also report the average number of entities and concepts for each dataset in Table \ref{table:datasets}.
All four datasets are tokenized through the jieba tool\footnote{https://github.com/fxsjy/jieba}.

\subsection{Compared Methods}
We compare our proposed model STCKA with the following methods:
\begin{itemize}
	\item CNN \cite{Kim2014Convolutional}: This model is a classic baseline for text classification. It uses CNN based on the pre-trained word embedding.
	
	\item RCNN \cite{lai2015recurrent}: This method uses a recurrent convolutional neural network for text classification. It applies RNN to capture contextual information and CNN to capture the key components in texts.
	
	\item CharCNN \cite{zhang2015character}. This method uses CNN with only character level features as the input. 
	
	\item BiLSTM-MP \cite{Lee2016Sequential}: This model is proposed for sequential short text classification. It uses a LSTM in each direction, and use max-pooling across all LSTM hidden states to get the sentence representation, then use a multi-layer perceptron to output the classification result.
	
	\item BiLSTM-SA \cite{Lin2017A}: This method uses BiLSTM and source2token self-attention to encode a sentence into a fixed size representation which is used for classification. 
	
	\item KPCNN \cite{Wang2017Combining}: This model is the state-of-the-art method for short text classification. It utilizes CNN to perform classification based on word and character level information of short text and concepts.
\end{itemize}

\subsection{Settings and Metrics}
For \textit{all models}, we use Adam \cite{kingma2014adam} for learning, with a learning rate of 0.01. The batch size is set to 64. The training epochs are set to 20. We use 50-dimension skip-gram character and word embedding \cite{Mikolov2013Efficient} pre-trained on Sogou News corpus\footnote{http://www.sogou.com/labs/resource/list\_news.php}. 
If a word is unknown, we will randomly initialize its embedding. 
We also use 50-dimension concept embedding which is randomly initialized. All character embedding, word embedding and concept embedding are trainable and fine-tuned in the training stage, since we hope to learn task-oriented representation. We use 1D CNN with filters of width [2,3,4] of size 50 for a total of 150.

For \textit{our model}, the following hyper-parameters are estimated based on the validation set and used in the final test set: $ u=64, d_a=70, d_b=35$. And 
$\gamma$ is automatically learned by the neural network, because this method achieves better classification results than using a fixed hyper-parameter.
The evaluation metric is \textit{accuracy}, which is widely used in text classification tasks \cite{Lee2016Sequential,Wang2017Combining}.

\subsection{Results}
We compare our model STCKA with six strong baselines and the results are shown in Table \ref{table:models}.
Our model outperforms traditional Deep Neural Networks (DNNs), including CNN, RCNN, CharCNN, BiLSTM-MP and BiLSTM-SA, without using any knowledge.
The main reason is that our model enriches the information of short texts with the help of KBs. Specifically, we incorporate prior knowledge in KBs into DNNs as explicit features which have a great contribution to short text classification. Compared with traditional DNNs, our model is more like a human being who has intrinsic ability to make decisions based on observation (i.e., training data for machines) as well as existing knowledge. 
In addition, our model also performs better than KPCNN since our model is able to pay more attention to important knowledge due to the attention mechanism. We use C-ST and C-CS attention to measure the importance of knowledge from two aspects and adaptively assign a proper weight to each knowledge of different short texts. 

\begin{table*}[!hbt]
	\center
	\begin{tabular}{|p{3cm}|p{1.5cm}|p{1.5cm}|p{2.5cm}|p{2cm}|}
		\hline
		Model&Weibo&Topic&Product Review&News Title\\
		\hline
		STCKA($\lambda=0.00$)&0.4280&0.8600&0.7390&0.7972 \\ 
		\hline
		STCKA($\lambda=0.25$)&\textbf{0.4320}&0.8700&\textbf{0.7430}&\textbf{0.8007} \\
		\hline
		STCKA($\lambda=0.50$)&0.4260&\textbf{0.8786}&0.7380&0.8002 \\
		\hline
		STCKA($\lambda=0.75$)&0.4220&0.8643&0.7380&0.7959 \\
		\hline
		STCKA($\lambda=1.00$)&0.4160&0.8557&0.7360&0.7965 \\
		\hline
	\end{tabular}
	\caption{The setting of hyper-parameter $\lambda$ on our model}
	\label{table:hyperparameter}
\end{table*}

\subsection{Knowledge Attention}\label{eotam}
The goal of this part is to verify the effectiveness of two attention mechanisms (i.e., C-ST and C-CS attention).
We manually tune the hyper-parameter $\gamma$ to explore the relative importance of C-ST and C-CS attention.
We vary $\gamma$ from 0 to 1 with an interval of 0.25, and the results are shown in 
Table \ref{table:hyperparameter}. In general, the model with $\gamma=0.25$ works better, but the advantage is not always there for different datasets. For instance, the model with $\gamma=0.50$ performs the best on Topic dataset. 
When $\gamma$ is equal to 0 or 1, the model performs poorly on all four datasets. 
Using C-ST attention only ($\gamma=1.00$), the model neglects the relative importance of each concept, which leads to poor performance. On the other hand, merely using C-CS attention ($\gamma=0.00$), the model ignores the semantic similarity between the short text and concepts. In this case, an improper concept may be assigned with a larger weight which also results in poor performance.

\begin{figure}[htbp]
	\centering
	\subfloat[The lable of the short text is \textit{fashion}. \label{fig:case1}] 
	{
		\includegraphics[width=.95\columnwidth]{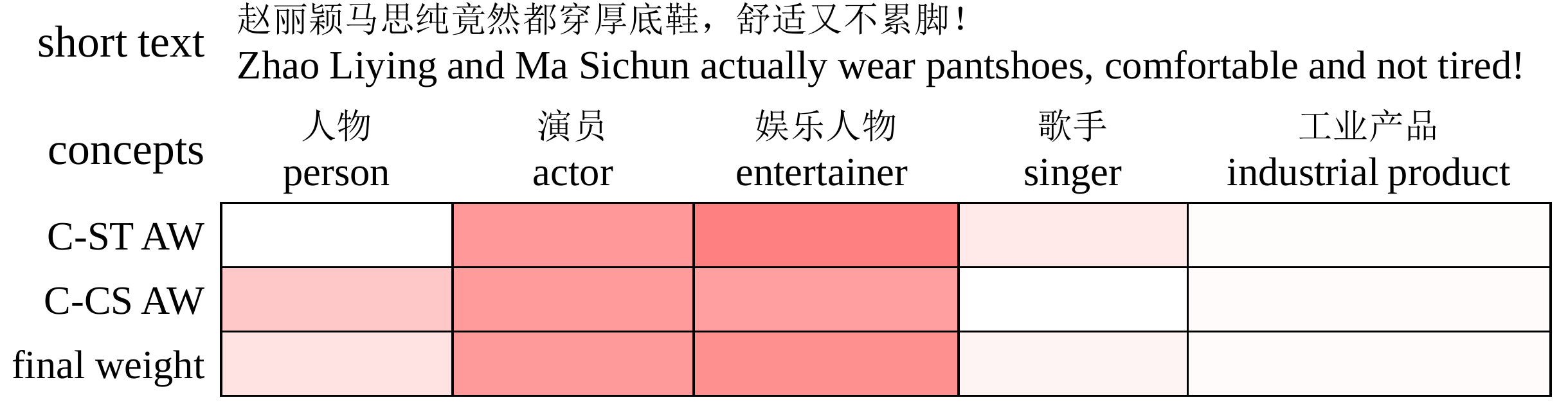}
	}\hfill
	\subfloat[The lable of the short text is \textit{car}.\label{fig:case2}]
	{
		\includegraphics[width=.95\columnwidth]{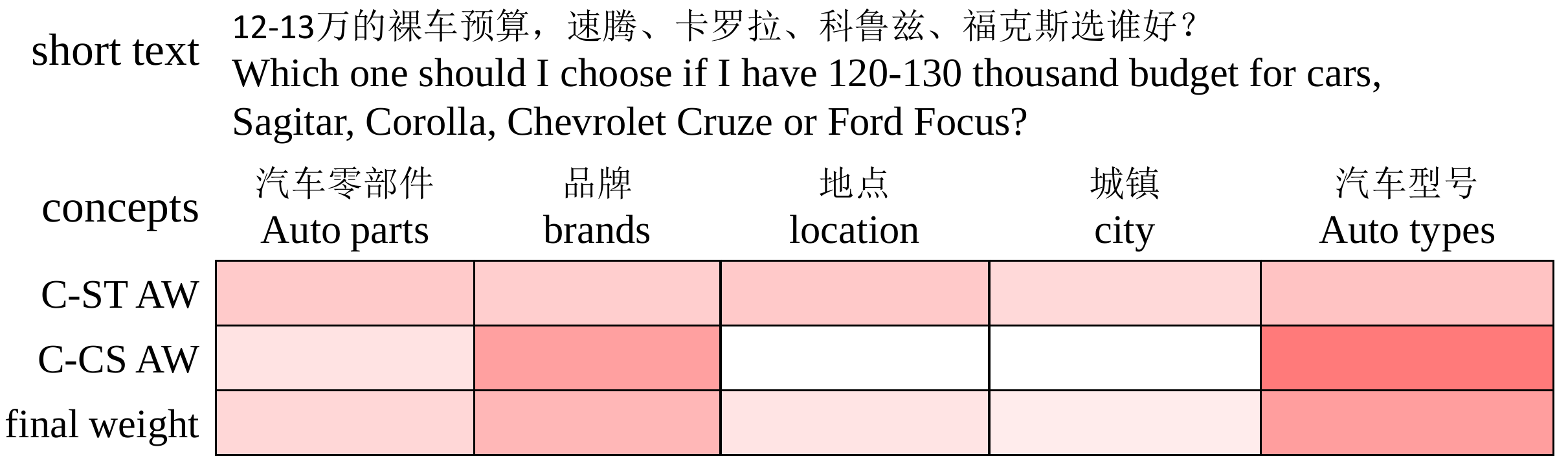}
	}\hfill
	\caption{Knowledge attention visualization. Attention Weight (AW) is used as the color-coding.} 
	\label{fig:case}
\end{figure}

To check whether the attention results conform to our intuition, we also pick some testing examples from the test set of News Title datasets and visualize their attention results in Figure \ref{fig:case}.
In general, an important concept for classification is assigned with a large weight and vice versa.
We also discover some characteristics of our model.
First, it is \textit{interpretable}. Given a short text and its corresponding concepts, our model tells us the contribution of each concept for classification by attention mechanism.
Second, it is \textit{robust} to the noisy concepts. For example, as shown in Figure \ref{fig:case1}, when conceptualizing the short text, we acquire some improper concepts such as \texttt{industrial product} which are not helpful for classification. Our model assigns a small attention weight to these concepts since they are irrelevant to the short text and have little similarity to the short text.
Third, the effect of C-CS attention is similar to that of \textit{feature selection}. To some extent, C-CS attention is a ``soft'' feature selection assigning a small weight (nearly to zero) to irrelevant concepts. Therefore, the solution (attention weight) produced by C-CS attention is sparse, which is similar to \textit{L1 Norm Regularization} \cite{park2007l1}. 

\subsection{Power of Knowledge}\label{cs}
We use conceptual information as prior knowledge to enrich the representation of short text and improve the performance of classification. 
The average number of entities and concepts of each dataset are shown in Table \ref{table:datasets}.
To verify the power of knowledge in our model, we pick some testing examples from Topic dataset and illustrate them in Table \ref{table:case3}.
These short texts are correctly classified by our model but misclassified by traditional DNNs that do not use any knowledge. In general, the conceptual information plays a crucial role in short text classification, especially when the context of short texts is not enough. 
As the first example shown in Table \ref{table:case3}, \texttt{Revolution of 1911} is a rare word, i.e., occurs less frequently in the training set, and thus is difficult to learn a good representation, resulting in poor performance of traditional DNNs.  
However, our model solves the \textit{rare and unknown word problem} \cite{gulcehre2016pointing} in some degree by introducing knowledge from KB. The concepts such as \texttt{history} and \texttt{historical event} used in our model are helpful for classifying the short text into the correct class \textit{history}.

\begin{figure}[!hbt]
	\centering
	\includegraphics[width=.95\columnwidth]{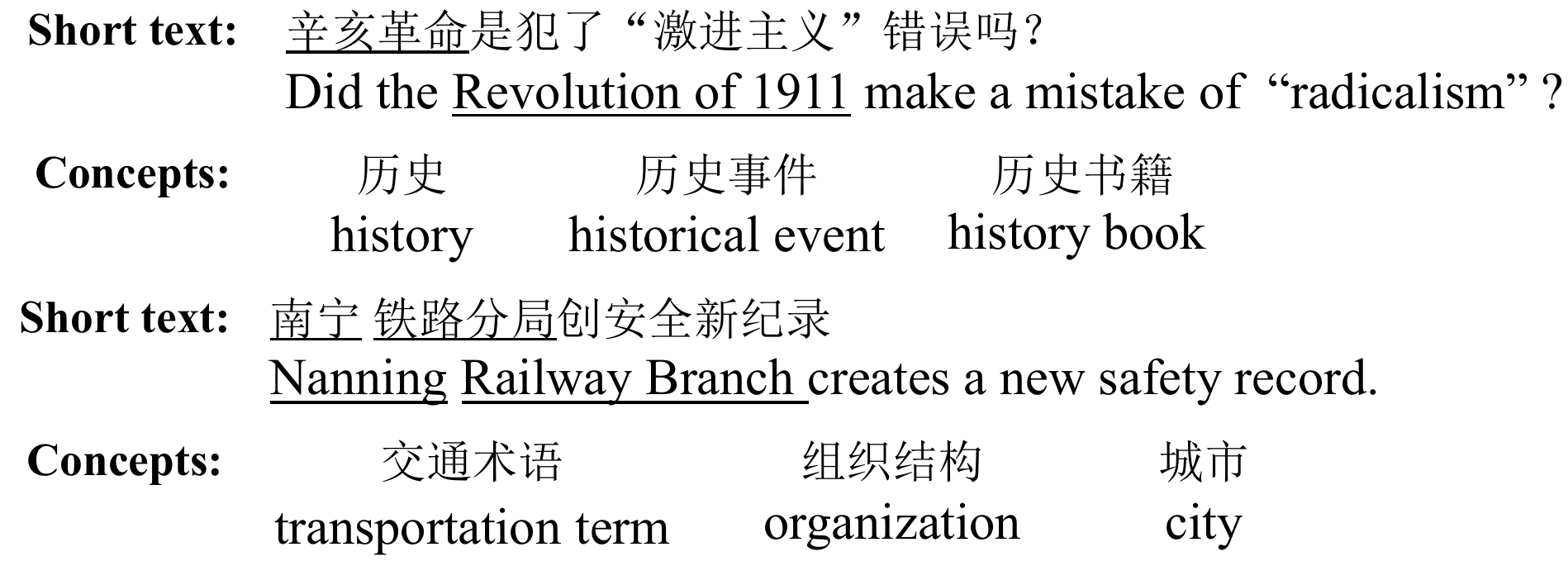}
	\caption{Two examples for power of knowledge. Underlined phrases are the entities, and the class labels of these two short texts are \textit{history} and \textit{transport} respectively.}
	\label{fig:case3}
\end{figure}

\begin{table*}[!hbt]
	\center
	\begin{tabular}{|p{3cm}|p{1.5cm}|p{1.5cm}|p{2.5cm}|p{2cm}|}
		\hline
		Model&Weibo&Topic&Product Review&News Title\\
		\hline
		STCKA-rand&0.3780&0.8414&0.7290&0.7930 \\ 
		\hline
		STCKA-static&0.4240&0.8600&0.7350&0.7889 \\
		\hline
		STCKA-non-static&\textbf{0.4320}&\textbf{0.8814}&\textbf{0.7430}&\textbf{0.8011} \\
		\hline
	\end{tabular}
	\caption{The impact of different embedding tunning methods on our model}
	\label{table:embedding}
\end{table*}

\subsection{Embedding Tunning}
We totally use three embeddings in our model. Concept embeddings are randomly initialized and fine-tuned in the training stage. As for character and word embedding, we try three embedding tuning strategies:
\begin{itemize}
	\item STCKA-rand: The embedding is randomly initialized and then modified in the training stage.
	\item STCKA-static: Using pre-trained embedding which is kept static in the training.
	\item STCKA-non-static: Using pre-trained embedding initially, and tuning it in the training stage.
\end{itemize}

As shown in Table \ref{table:embedding}, in general, STCKA-non-static performs the best on all four datasets since it makes full use of pre-trained word embedding and fine-tunes it during training phrase to capture specific information on different tasks. 
Besides, STCKA-rand performs more poorly than STCKA-static on small training datasets such as Weibo and Topic. The reason could be twofold: (1) The amount of labeled samples in the two experimental datasets is too small to tune reliable embeddings from scratch for the in-vocabulary words (i.e., existing in the training data); (2) A lot of out-of-vocabulary words, i.e., absent from the training data, but exist in the testing data. 
However, STCKA-rand outperforms STCKA-static on large-scale training data such as News Title. 
Because large-scale training data alleviates two above-mentioned reasons and enables STCKA-rand to learn task-oriented embeddings which are better for different classification tasks.

\subsection{Error Analysis}
We analyze the bad cases induced by our proposed model on News Title dataset. Most of the bad cases can be
generalized into two categories. 
First, long-tailed entities lack discriminative knowledge in KB due to the incompleteness of KB. For example, in short text ``\textit{what does a radio mean to sentry in the cold night alone}'', the entity \texttt{sentry} is a long-tailed entity without useful concepts in KB. Thus, the short text cannot be classified into the correct class \textit{military}.
Second, some short texts are too short and lack contextual information. Even worse, there are no entities mentioned in these short texts which leads to the failure of conceptualization. Therefore, it is difficult to classify the short text ``\textit{don't pay money, it's all routines}'' into the class \textit{fashion}.

\section{Related Works}
\textbf{Short Text Classification} Existing methods for text classification can be divided into two categories: explicit representation and implicit representation.
The explicit model depends on human-designed features and represents a short text as a sparse vector.
\cite{cavnar1994n} made full use of simple n-gram features for text classification.
\cite{Pang2002Thumbs,Post2013Explicit} exploited more complex features such as POS tagging and dependency parsing to improve the performance of classification.
Some researches introduced knowledge from KBs to enrich the information of short texts. \cite{Gabrilovich2007Computing} utilized the Wikipedia information to enrich the text representation. \cite{Wang2014Concept} conceptualized a short text to a set of relevant concepts which are used for classification by leveraging Probase. 
The explicit model is interpretable and easily understood by human beings, but neglects the context of short texts and cannot capture deep semantic information.

Recently, implicit models are widely used in text classification due to the development of deep learning.
The implicit model maps a short text to an implicit space and represents it as a dense vector.  
\cite{Kim2014Convolutional} used CNN with pre-trained word vectors for sentence classification. 
\cite{lai2015recurrent} presented a model based on RNN and CNN for short text classification.
\cite{zhang2015character} offered an empirical exploration on the use of character-level CNN for text classification.
\cite{Lee2016Sequential} proposed a RNN model for sequential short text classification. It used BiLSTM and max-pooling across all LSTM hidden states to produce the representation of short text.
\cite{Lin2017A} classified the texts by relying on BiLSTM and self-attention.
The implicit model is good at capturing syntax and semantic information in short text, but ignores the important prior knowledge that can be acquired from KBs.
\cite{Wang2017Combining} introduced knowledge from Probase into deep neural networks to enrich the representation of short text.
However, Wang et al.'s work has two limitations: 1) failing to consider the semantic similarity between short texts and knowledge; 2) ignoring the relative importance of each knowledge. 

\textbf{Attention Mechanism} has been successfully used in many NLP tasks.
According to the attention target, it can be divided into vanilla attention and self-attention. \cite{Bahdanau2015Neural} first used vanilla attention to compute the attention score between a query and each input token in machine translation task. \cite{Hao2017An} employed vanilla attention to measure the similarity between question and answer in question answering task. 
Self-attention can be divided into two categories including token2token self-attention and source2token self-attention. \cite{Vaswani2017Attention} applied token2token self-attention to neural machine translation and achieved the state-of-the-art performance. \cite{Lin2017A} used source2token self-attention to explore the importance of each token to the entire sentences in sentence representation task. 
Inspired by these work, we employ two attention mechanisms from two aspects to measure the importance of knowledge.

\section{Conclusion and Future work}
In this paper, we propose deep Short Text Classification with Knowledge
Powered Attention. We integrate the conceptual information in KBs to enhance the representation of short text. To measure the importance of each concept, we apply two attention mechanisms to automatically acquire the weight of concepts that is used for generating the concept representation. We classify a short text based on the text and its relevant concepts. Finally, we demonstrate the effectiveness of our model on four datasets for different tasks, and the results show that it outperforms the state-of-the-art methods.

In the future, we will incorporate \textit{property-value information} into deep neural networks to further improve the performance of short text classification.
We find that some entities mentioned in short texts lack concepts due to the incompleteness of KB. Apart from conceptual information, entity properties and their values can also be injected into deep neural networks as explicit features. For example, the entity \texttt{Aircraft Carrier} has a property-value pair \texttt{domain}-\texttt{military}, which is an effective feature for classification.

\bibliographystyle{named}
\bibliography{AAAI-ChenJ.4251}

\end{document}